\documentclass[journal]{IEEEtran}
\usepackage{multirow}
\usepackage{amssymb,amsmath}
\usepackage{rotating}
\usepackage[ruled,vlined]{algorithm2e}
\usepackage{algorithmic}
\usepackage{bm}
\usepackage{graphicx}
\usepackage{subfigure}
\usepackage{color}
\usepackage{amsthm}
\usepackage{amsbsy}

\usepackage{tabularx}
\usepackage{multirow}
\newcommand{\tabincell}[2]{\begin{tabular}{@{}#1@{}}#2\end{tabular}}
\newcommand{\ie}{\textit{i}.\textit{e}.}
\newcommand{\eg}{\textit{e}.\textit{g}.}

\ifCLASSINFOpdf
\else
\fi
\begin{document}
%
\title{Robust Face Recognition via Multimodal Deep \\Face Representation}
%
%

\author{Changxing~Ding,~\IEEEmembership{Student~Member,~IEEE},
        Dacheng~Tao,~\IEEEmembership{Fellow,~IEEE}
\thanks{C. Ding and D. Tao are with the Centre for Quantum Computation and Intelligent Systems,
and the Faculty of Engineering and Information Technology, University of Technology, Sydney,
81 Broadway, Ultimo, NSW 2007, Australia
(email: changxing.ding@student.uts.edu.au, dacheng.tao@uts.edu.au).}}

%
%

\markboth{To Appear in IEEE TRANSACTIONS ON MULTIMEDIA,~2015}%
{Ding \MakeLowercase{\textit{et al.}}: Robust Face Recognition via Multimodal Deep Face Representation}
%



\maketitle

\begin{abstract}
Face images appeared in multimedia applications, \eg, social networks and digital entertainment, usually exhibit dramatic pose, illumination, and expression variations,
resulting in considerable performance degradation for traditional face recognition algorithms.
This paper proposes a comprehensive deep learning framework to jointly learn face representation using multimodal information.
The proposed deep learning structure is composed of a set of elaborately designed convolutional neural networks (CNNs) and a three-layer stacked auto-encoder (SAE).
The set of CNNs extracts complementary facial features from multimodal data.
Then, the extracted features are concatenated to form a high-dimensional feature vector,
whose dimension is compressed by SAE.
All the CNNs are trained using a subset of 9,000 subjects from the publicly available CASIA-WebFace database,
which ensures the reproducibility of this work.
Using the proposed single CNN architecture and limited training data, 98.43\% verification rate is achieved on the LFW database.
Benefited from the complementary information contained in multimodal data,
our small ensemble system achieves higher than 99.0\% recognition rate on LFW using publicly available training set.
\end{abstract}


%
\IEEEpeerreviewmaketitle

\section{Introduction}
\IEEEPARstart{F}{ace} recognition has been one of the most extensively studied topics in computer vision.
The importance of face recognition is closely related to its great potential in multimedia applications,
\eg, photo album management in social networks, human machine interaction, and digital entertainment.
With years of effort, significant progress has been achieved for face recognition.
However, it remains a challenging task for multimedia applications, as observed in recent works~\cite{ding2015comprehensive,choi2011collaborative}.
In this paper, we handle the face recognition problem for matching internet face images appeared in social networks,
which is one of the most common applications in multimedia circumstances.

\begin{figure}
\centering
\includegraphics[width=1.0\linewidth]{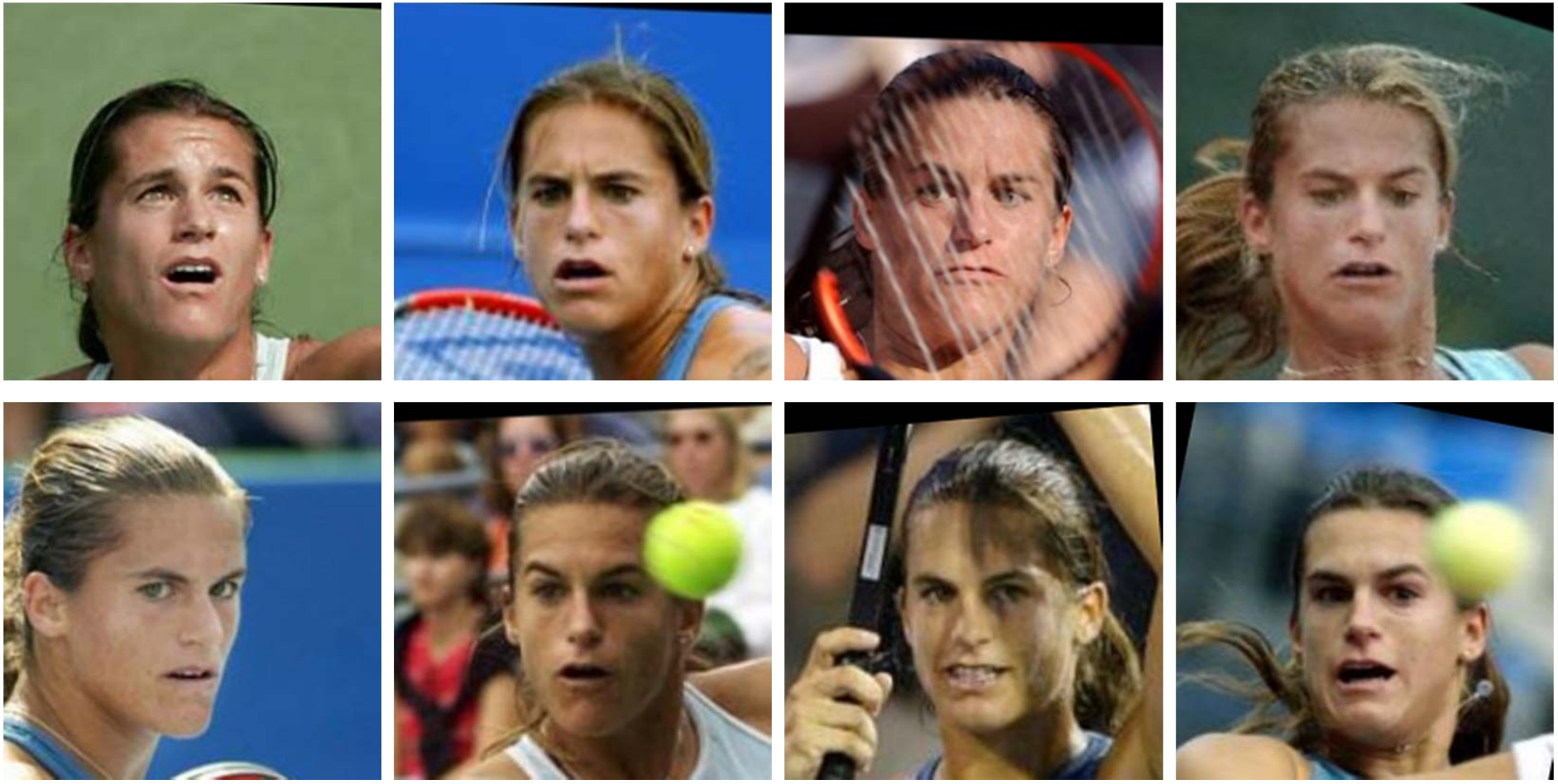}
\caption{Face images in multimedia applications usually exhibit rich variations in pose, illumination, expression, and occlusion.}
\label{fig:lfwImages}
\end{figure}

Recognizing the face images appeared in social networks is difficult, due to the reasons mainly from the following two perspectives.
First, the face images uploaded to social networks are captured in real-world conditions;
therefore faces in these images usually exhibit rich variations in pose, illumination, expression, and occlusion, as illustrated in Fig.~\ref{fig:lfwImages}.
Second, face recognition in social networks is a large-scale recognition problem due to the numerous face images of potentially large amount of users.
The prediction accuracy of face recognition algorithms usually degrades dramatically with the increase of face identities.

Accurate face recognition depends on high quality face representations.
Good face representation should be discriminative to the change of face identify while remains robust to intra-personal variations.
Conventional face representations are built on local descriptors,
\eg, Local Binary Patterns (LBP)~\cite{ahonen2006face}, Local Phase Quantization (LPQ)~\cite{ojansivu2008blur,ahonen2008recognition},
Dual-Cross Patterns (DCP)~\cite{ding2014multi}, and Binarised Statistical Image Features (BSIF)~\cite{kannala2012bsif}.
However, the representation composed by local descriptors is too shallow to differentiate the complex nonlinear facial appearance variations.
To handle this problem, recent works turn to Convolutional Neural Networks (CNNs)~\cite{taigman2014deepface,sun2014deep} to automatically learn effective features
that are robust to the nonlinear appearance variation of face images.
However, the existing works of CNN on face recognition extract features from limited modalities,
the complementary information contained in more modalities is not well studied.

Inspired by the complementary information contained in multi-modalities and the recent progress of deep learning on various fields of computer vision,
we present a novel face representation framework that adopts an ensemble of CNNs to leverage the multimodal information.
The performance of the proposed multimodal system is optimized from two perspectives.
First, the architecture for single CNN is elaborately designed and optimized with extensive experimentations.
Second, a set of CNNs is designed to extract complementary information from multiple modalities,
\ie, the holistic face image, the rendered frontal face image by 3D model, and uniformly sampled face patches.
Besides, we design different structures for different modalities, \ie,
a complex structure is designed for the modality that contains the richest information
while a simple structure is proposed for the modalities with less information.
In this way, we strike a balance between recognition performance and efficiency.
The capacity of each modality for face recognition is also compared and discussed.

\begin{figure*}
\centering
\includegraphics[width=0.7\linewidth]{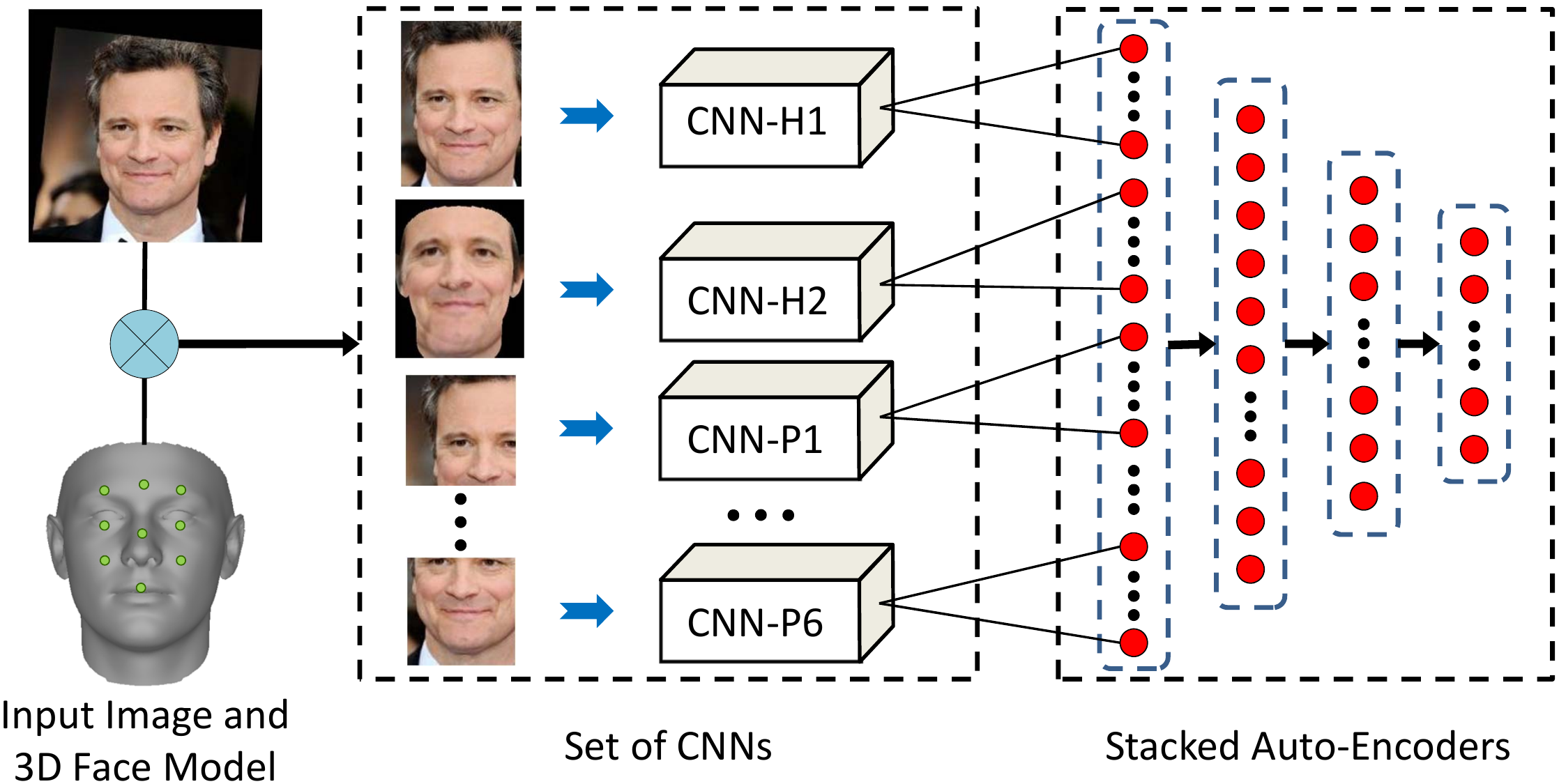}
\caption{Flowchart of the proposed multimodal deep face representation (MM-DFR) framework.
MM-DFR is essentially composed of two steps: multimodal feature extraction using a set of CNNs,
and feature-level fusion of the set of CNN features using SAE.}
\label{fig:MMDFR_Framework}
\end{figure*}

We term the proposed deep learning-based face representation scheme as Multimodal Deep Face Representation (MM-DFR), as illustrated in Fig.~\ref{fig:MMDFR_Framework}.
Under this framework, the face representation of one face image involves feature extraction using each of the designed CNNs.
The extracted features are concatenated as the raw feature vector, whose dimension is compressed by a three-layer SAE.
Extensive experiments on the Labeled Face in the Wild (LFW)~\cite{LFWTech} and CASIA-WebFace databases~\cite{yi2014learning}
indicate that superior performance is achieved with the proposed MM-DFR framework.
Besides, the influence of several implementation details, \eg,
the usage strategies of ReLU nonlinearity, multiple modalities, aggressive data augmentation,
multi-stage training, and L2 normalization, is compared and discussed in the experimentation section.
To the best of our knowledge, this is the first published approach that achieves higher than 99.0\% recognition rate using a publicly available training set on the LFW database.

The remainder of the paper is organized as follows:
Section II briefly reviews related works for face recognition and deep learning.
The proposed MM-DFR face representation scheme is illustrated in Section III.
Face matching using MM-DFR is described in Section IV.
Experimental results are presented in Section V,
leading to conclusions in Section VI.

\section{Related Studies}
\subsection{Face Image Representation}
Popular face representations can be broadly grouped into two categories:
local descriptor-based representations and deep learning-based representations.

Traditional face representations are based on local descriptors~\cite{gunther20132013,Ross2015report}.
Local descriptors can be further divided into two groups: the handcrafted descriptors and the learning-based descriptors.
Among the handcrafted descriptors, Ahonen \textit{et al.}~\cite{ahonen2006face} proposed to employ the texture descriptor LBP for face representation.
LBP works by encoding the gray-value difference between each pixel and its neighboring pixels into binary codes.
Ding \textit{et al.}~\cite{ding2014multi} proposed the Dual-Cross Patterns (DCP) descriptor
to encode second order statistics along the distribution directions of facial components.
Other effective handcrafted local descriptors include Local Phase Quantization (LPQ)~\cite{ojansivu2008blur} and Gabor-based descriptors.
Representative learning-based descriptors include Binarised Statistical Image Features (BSIF)~\cite{kannala2012bsif,rahimzadeh2014dynamic}
and Discriminant Face Descriptor (DFD)~\cite{lei2014learning}, \textit{et al.}.
Compared with the handcrafted descriptors, the learning-based descriptors usually optimize the pattern encoding step using machine learning techniques.
An extensive and systematic comparison among existing local descriptors for face recognition can be found in~\cite{ding2014multi};
and a detailed summarization on local descriptor-based face representations can be found in a recent survey~\cite{ding2015comprehensive}.
Despite of its ease of use, the local descriptor-based approaches have clear limitations:
the constructed face reprsentation is sensitive to the non-linear intra-personal variations,
\eg, pose~\cite{ding2015multi}, expression~\cite{hsieh2009expression}, and illumination~\cite{Ross2015report}.
In particular, the intra-personal appearance change caused by pose variations may substantially surpass the difference caused by identities~\cite{ding2015multi}.

The complicated facial appearance variations call for non-linear techniques for robust face representation,
and recent progress on deep learning provides an effective tool.
In the following, we review the most relevant progress for deep learning-based face recognition.
Taigman \textit{et al.}~\cite{taigman2014deepface} proposed the DeepFace architecture for face recognition.
They use the softmax loss, \ie, the face identification loss, as the supervisory signal to train the network
and achieve high recognition accuracy approaching the human-level.
Sun \textit{et al.}~\cite{sun2014deep} proposed to combine the identification and verification losses for more effective training.
They empirically verified that the combined supervisory signal is helpful to promote the discriminative power of extracted CNN features.
Zhou \textit{et al.}~\cite{zhou2015naive} investigated the influence of distribution and size of training data to the performance of CNN.
With a huge training set composed of 5 millions of labelled faces, they achieved an accuracy of 99.5\% accuracy on LFW using naive CNN structures.
One common problem for the above works is that they all employ private face databases for training.
Due to the distinct size and unknown distribution of these private data,
the performance of the above works may not be directly comparable.
Recently, Yi \textit{et al.}~\cite{yi2014learning} released the CASIA-WebFace database which contains 494,414 labeled images of 10,575 subjects.
The availability of such a large-scale database enables researchers to compete on a fair starting line.
In this paper, the training of all CNNs are conducted exclusively on a subset of 9,000 subjects of the CASIA-WebFace database,
which ensures the reproducibility of this work.
The CNN architectures designed in this paper are inspired by two previous works~\cite{simonyan2014very,yi2014learning},
but with a number of modifications and improvements,
and our designed CNN models have visible advantage in performance.

\subsection{Multimodal-based Face Recognition}
Most of face recognition algorithms extract a single face representation from the face image.
However, they are restrictive in capturing the diverse information contained in the face image.
To handle this problem, Ding \textit{et al.}~\cite{ding2014multi} proposed to extract the Multi-directional Multi-level DCPs (MDML-DCPs) feature
which includes three holistic-level features and six component-level features.
The set of the nine facial features composes the face representation.
Similar strategies have been adopted in deep learning-based face representations.
For example, the DeepFace approach~\cite{taigman2014deepface} adopts the same CNN structure to extract facial features from RGB image, gray-level image
and gradient map. The set of face representations are fused in the score level.
Sun \textit{et al.}~\cite{sun2014deep} proposed to extract deep features from 25 image patches cropped with various scales and positions.
The dimension of the concatenated deep features is reduced by Principle Component Analysis (PCA).
Multimodal systems that fuse multiple feature cues are also employed in other topics of multimedia and computer vision,
\eg, visual tracking~\cite{mei2015robust}, image classification~\cite{qi2009two,xu2014large,qi2008two},
and social media analysis~\cite{qi2009learning,eaton2014multi,qi2012exploring,hussain2015co,qi2007correlative}.

Our multimodal face recognition system is related to the previous approaches, and there is clear novelty.
First, we extract multimodal features from the holistic face image, rendered frontal face by 3D face model, and uniformly sampled image patches.
The three modalities stand for holistic facial features and local facial features, respectively.
Different from~\cite{taigman2014deepface} that employs the 3D model to assist 2D piece-wise face warping,
we utilize the 3D model to render a frontal face in 3D domain, which indicates much stronger alignment compared with~\cite{taigman2014deepface}.
Different from~\cite{sun2014deep} that randomly crops 25 patches over the face image using dense facial feature points,
we uniformly sample a small number of patches with the help of 3D model and sparse facial landmarks, which is more reliable compared with dense landmarks.
Second, we propose to employ SAE to compress the high-dimensional deep feature into a compact face signature.
Compared with the traditional PCA approach for dimension reduction, SAE has advantage in learning non-linear feature transformations.
Third, the large-scale unconstrained face identification problem has not been well studied due to the lack of appropriate face databases.
Fortunately, the recently published CASIA-WebFace~\cite{yi2014learning} database provides the possibility for such kind of evaluation.
In this paper, we evaluate the identification performance of MM-DFR on the CASIA-WebFace database.

\section{Multimodal Deep Face Representation}
In this section, we describe the proposed MM-DFR framework for face representation.
As shown in Fig.~\ref{fig:MMDFR_Framework}, MM-DFR is essentially composed of two steps:
multimodal feature extraction using a set of CNNs,
and feature-level fusion of the set of CNN features using SAE.
In the following, we describe the two main components in detail.

\subsection{Single CNN Architecture}
\begin{figure}
\centering
\includegraphics[width=1.0\linewidth]{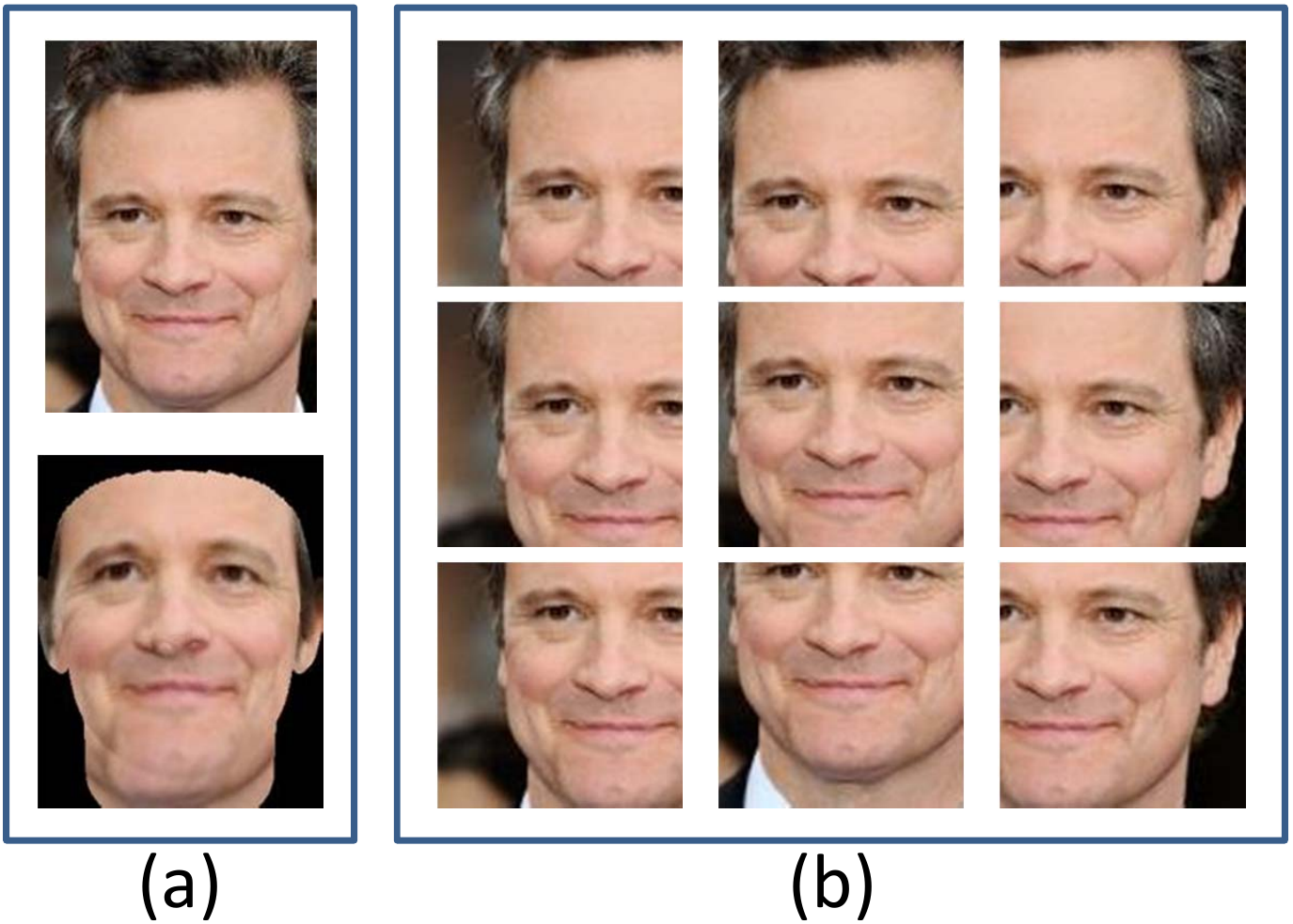}
\caption{The normalized holistic face images and image patches as input for MM-DFR.
(a) The original holistic face image and the 3D pose normalized holistic image;
(b) Image patches uniformly sampled from the original face image.
Due to facial symmetry and the augmentation by horizontal flipping, we only leverage the six patches illustrated in the first two columns.
}
\label{fig:CroppedIm}
\end{figure}

All face images employed in this paper are first normalized to $230\times230$ pixels with an affine transformation according to the coordinates of five sparse facial feature points,
\ie, both eye centers, the nose tip, and both mouth corners.
Sample images after the affine transformation are illustrated in Fig.~\ref{fig:lfwImages}.
We employ an off-the-shelf face alignment tool~\cite{zhang2014facial} for facial feature detection.
Based on the normalized image, one holistic face image of size $165\times120$ pixels (Fig.~\ref{fig:CroppedIm}a)
and six image patches of size $100\times100$ pixels (Fig.~\ref{fig:CroppedIm}b) are sampled.
Another holistic face image is obtained by 3D pose normalization using OpenGL~\cite{ding2015multi}.
Pose variation is reduced in the rendered frontal face, as shown in Fig.~\ref{fig:CroppedIm}a.

Two CNN models named NN1 and NN2 are designed, which are closely related to the ones proposed in~\cite{simonyan2014very,yi2014learning},
but with a number of modifications and improvements.
We denote the CNN that extracts feature from the holistic face image as CNN-H1.
In the following, we take CNN-H1 for example to illustrate the architectures of NN1 and NN2, as shown in Table~\ref{tab:NN1} and Table~\ref{tab:NN2}, respectively.
The other seven CNNs employ similar structure but with modifications in resolution for each layer.
The major difference between NN1 and NN2 is that NN2 is both deeper and wider than NN1.
With larger structure, NN2 is more robust to highly non-linear facial appearance variations; therefore, we apply it to CNN-H1.
NN1 is smaller but more efficient and we apply it to the other seven CNNs,
with the underlying assumption that the image patches and pose normalized face contain less nonlinear appearance variations.
Compared with NN1, NN2 is more vulnerable to overfitting due to its larger number of parameters.
In this paper, we make use of aggressive data augmentation and multi-stage training strategies to reduce overfitting.
Details of the two strategies are described in the experimentation section.

NN1 contains 10 convolutional layers, 4 max-pooling layers, 1 mean-pooling layer, and 2 fully-connected layers.
In comparison, NN2 incorporates 12 convolutional layers.
Small filters of $3\times3$ are utilized for all convolutional layers.
As argued in~\cite{simonyan2014very}, successive convolutions by small filters equal to one convolution operation by a large filter,
but effectively enhances the model's discriminative power and reduces the number of filter parameters to learn.
ReLU~\cite{dahl2013improving} activation function is utilized after all but the last convolutional layers.
The removal of ReLU nonlinearity helps to generate dense features, as described in~\cite{yi2014learning}.
We also remove the ReLU nonlinearity after Fc6;
therefore the projection of convolutional features by Fc6 layer is from dense to dense,
which means that Fc6 effectively equals to a linear dimension reduction layer that is similar to PCA or Linear Discriminative Analysis (LDA).
This is different from previous works that favor sparse features produced by ReLU~\cite{taigman2014deepface,sun2014deep,sun2015deeply}.
Our model is also different from~\cite{yi2014learning} since~\cite{yi2014learning} simply removes the linear dimension reduction layer (Fc6).
The output of the Fc6 layer is employed as face representation.
In the experimental section, we empirically justify that the dense-to-dense projection by Fc6 is advantageous to produce more discriminative features.
The forward function of ReLU is represented as
\begin{equation}
\begin{array}{cl}
R(x_{i})=max(0,W_{c}^{T}x_{i}+b_{c}),
\end{array}
\label{E:ReLU}
\end{equation}
where $x_{i}$, $W_{c}$, and $b_{c}$ are the input, weight, and bias of the corresponding convolutional layer before the ReLU activation function.
$R(x_{i})$ is the output of the ReLU activation function.
The dimension of the Fc6 layer is set to 512.
The dimension of the Fc7 is set to 9000, which equals to the number of training subjects employed in this paper.
We employ dropout~\cite{krizhevsky2012imagenet} as a regularizer on the first fully-connected layer in the case of overfitting caused by the large amount of parameters.
The dropout ratio is set to 0.4.
Since this low-dimensional face representation is utilized to distinguish as large as 9,000 subjects in the training set,
it should be very discriminative and has good generalization ability.

\begin{table}[!t]
\renewcommand{\arraystretch}{1.3}
\caption{Details of the model architecture for NN1}
\label{tab:NN1}
\centering
\begin{tabular}{|l|c|c|c|c|c|}
\hline
Name         &Type         &\tabincell{c}{Input\\Size}   &\tabincell{c}{Filter\\Number}  &\tabincell{c}{Filter Size\\/stride /pad}  &\tabincell{c}{With\\Relu}\\\hline\hline
Conv11       &conv         &165$\times$120     &64          &3$\times$3 /1 /0   &yes\\
Conv12       &conv         &163$\times$118     &128         &3$\times$3 /1 /0   &yes\\
Pool1        &max pool     &161$\times$116     &N/A         &2$\times$2 /2 /0   &no\\\hline\hline
Conv21       &conv         &80$\times$58       &64          &3$\times$3 /1 /0   &yes\\
Conv22       &conv         &78$\times$56       &128         &3$\times$3 /1 /0   &yes\\
Pool2        &max pool     &76$\times$54       &N/A         &2$\times$2 /2 /0   &no\\\hline\hline
Conv31       &conv         &38$\times$27       &64          &3$\times$3 /1 /1   &yes\\
Conv32       &conv         &38$\times$27       &128         &3$\times$3 /1 /1   &yes\\
Pool3        &max pool     &38$\times$27       &N/A         &2$\times$2 /2 /1   &no\\\hline\hline
Conv41       &conv         &20$\times$14       &128         &3$\times$3 /1 /1   &yes\\
Conv42       &conv         &20$\times$14       &256         &3$\times$3 /1 /1   &yes\\
Pool4        &max pool     &20$\times$14       &N/A         &2$\times$2 /2 /0   &no\\\hline\hline
Conv51       &conv         &10$\times$7        &128         &3$\times$3 /1 /1   &yes\\
Conv52       &conv         &10$\times$7        &256         &3$\times$3 /1 /1   &\textbf{no}\\
Pool5        &mean pool    &10$\times$7        &N/A         &2$\times$2 /2 /1   &no\\\hline\hline
Dropout      &dropout      &6144$\times$1       &N/A         &N/A             &N/A \\
Fc6          &fully-conn   &512$\times$1        &N/A         &N/A             &\textbf{no}\\
Fc7          &fully-conn   &9000$\times$1       &N/A         &N/A             &no\\
Softmax      &softmax      &9000$\times$1       &N/A         &N/A             &N/A\\\hline
\end{tabular}
\end{table}

\begin{table}[!t]
\renewcommand{\arraystretch}{1.3}
\caption{Details of the model architecture for NN2}
\label{tab:NN2}
\centering
\begin{tabular}{|l|c|c|c|c|c|}
\hline
Name         &Type         &\tabincell{c}{Input\\Size}   &\tabincell{c}{Filter\\Number}  &\tabincell{c}{Filter Size\\/stride /pad}  &\tabincell{c}{With\\Relu}\\\hline\hline
Conv11       &conv         &165$\times$120     &64          &3$\times$3 /1 /0   &yes\\
Conv12       &conv         &163$\times$118     &128         &3$\times$3 /1 /0   &yes\\
Pool1        &max pool     &161$\times$116     &N/A         &2$\times$2 /2 /0   &no\\\hline\hline
Conv21       &conv         &80$\times$58       &64          &3$\times$3 /1 /0   &yes\\
Conv22       &conv         &78$\times$56       &128         &3$\times$3 /1 /0   &yes\\
Pool2        &max pool     &76$\times$54       &N/A         &2$\times$2 /2 /0   &no\\\hline\hline
Conv31       &conv         &38$\times$27       &128          &3$\times$3 /1 /1   &yes\\
Conv32       &conv         &38$\times$27       &128         &3$\times$3 /1 /1   &yes\\
Pool3        &max pool     &38$\times$27       &N/A         &2$\times$2 /2 /1   &no\\\hline\hline
Conv41       &conv         &20$\times$14       &256         &3$\times$3 /1 /1   &yes\\
Conv42       &conv         &20$\times$14       &256         &3$\times$3 /1 /1   &yes\\
Conv43       &conv         &20$\times$14       &256         &3$\times$3 /1 /1   &yes\\
Pool4        &max pool     &20$\times$14       &N/A         &2$\times$2 /2 /0   &no\\\hline\hline
Conv51       &conv         &10$\times$7        &256         &3$\times$3 /1 /1   &yes\\
Conv52       &conv         &10$\times$7        &256         &3$\times$3 /1 /1   &yes\\
Conv53       &conv         &10$\times$7        &256         &3$\times$3 /1 /1   &\textbf{no}\\
Pool5        &mean pool    &10$\times$7        &N/A         &2$\times$2 /2 /1   &no\\\hline\hline
Dropout      &dropout      &6144$\times$1       &N/A         &N/A             &N/A \\
Fc6          &fully-conn   &512$\times$1        &N/A         &N/A             &\textbf{no}\\
Fc7          &fully-conn   &9000$\times$1       &N/A         &N/A             &no\\
Softmax      &softmax      &9000$\times$1       &N/A         &N/A             &N/A\\\hline
\end{tabular}
\end{table}

The other holistic image is rendered by OpenGL with the help of 3D generic face model~\cite{ding2015multi}.
Pose variation is reduced in the rendered image.
We denote the CNN that extracts deep feature from this image as CNN-H2, as illustrated in Fig.~\ref{fig:MMDFR_Framework}.
Therefore, the first two CNNs encode holistic image features from different modalities.
The CNNs that extract features from the six image patches are denoted as CNN-P1, CNN-P2, to CNN-P6, respectively, as illustrated in Fig.~\ref{fig:MMDFR_Framework}.
Exactly the same network structure is adopted for each of the six CNNs.

\begin{figure}
\centering
\includegraphics[width=0.95\linewidth]{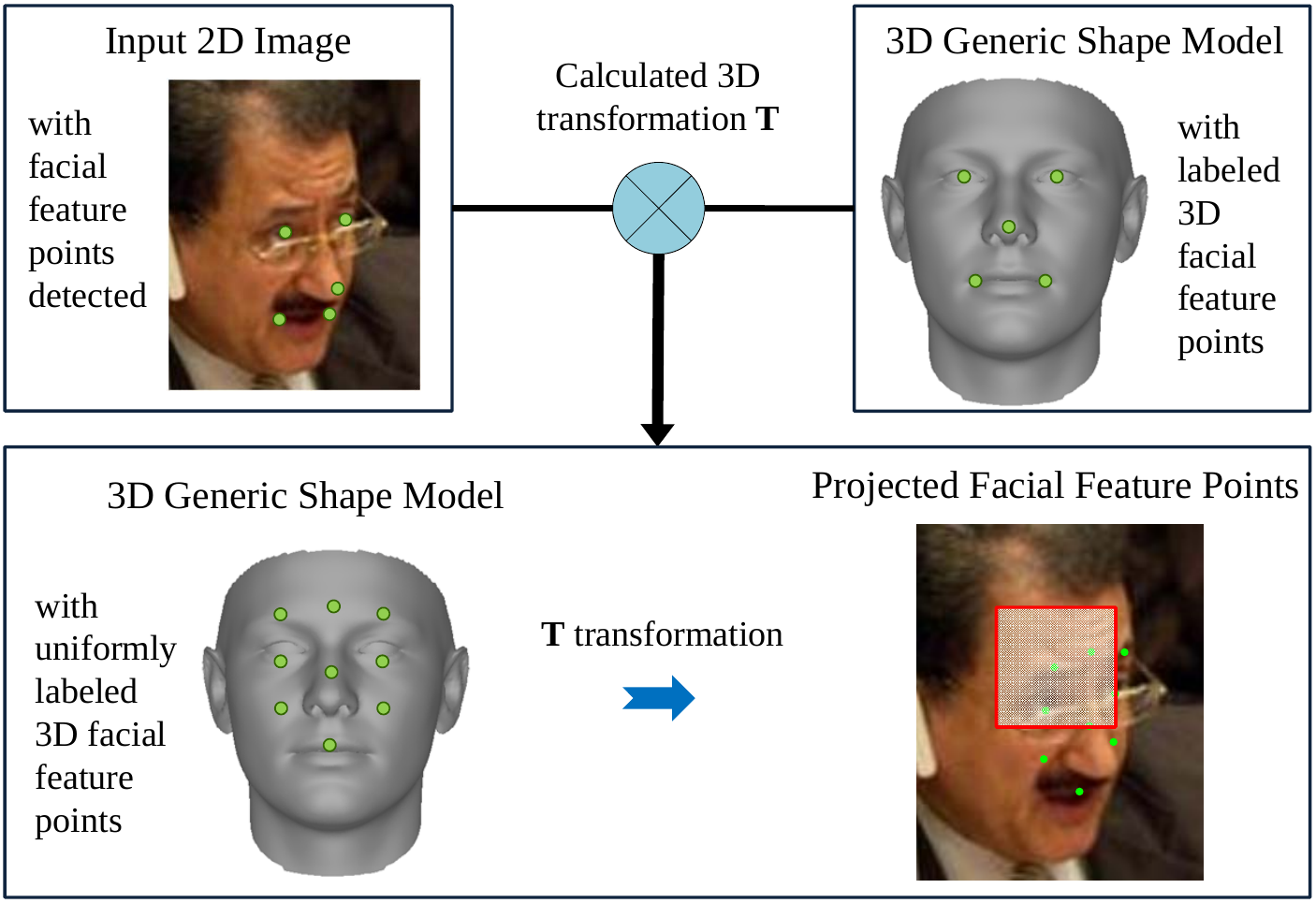}
\caption{The principle of patch sampling adopted in this paper.
A set of 3D landmarks are uniformly labeled on the 3D face model,
and are projected to the 2D image.
Centering around each landmark, a square patch of size $100\times100$ pixels is cropped, as illustrated in Fig.~\ref{fig:CroppedIm}b.}
\label{fig:ThreeDProjectionPrinciple}
\end{figure}

\begin{figure}
\centering
\includegraphics[width=1.0\linewidth]{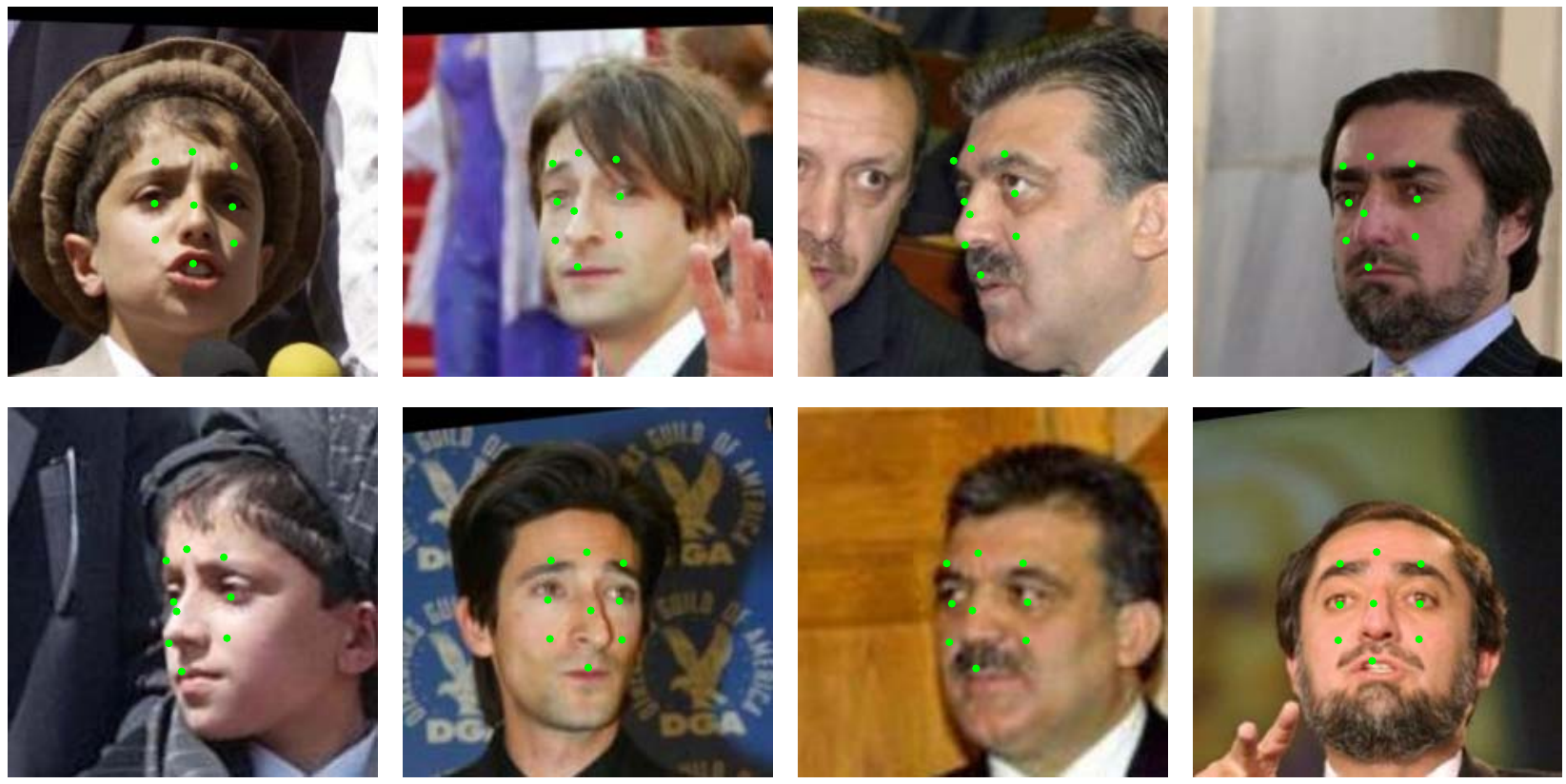}
\caption{More examples about the uniformly detected landmarks that are projected from a generic 3D face model to 2D images.}
\label{fig:ThreeDProjectedLandmarks}
\end{figure}

Different from previous works that randomly sample a large number of image patches~\cite{sun2014deep},
we propose to sample a small number of image patches uniformly in the semantic meaning,
which contributes to maximizing the complementary information contained within the sampled patches.
However, the uniform sampling of the image patches is not easy due to the pose variations of the face appeared in real-world images, as shown in Fig.~\ref{fig:lfwImages}.
We tackle this problem with a recently proposed strategy for pose-invariant face recognition~\cite{yi2013towards}.
The principle of the patch sampling process is illustrated in Fig.~\ref{fig:ThreeDProjectionPrinciple}.
In brief, nine 3D landmarks are manually labeled on a generic 3D face model and the 3D landmarks spread uniformly across the face model.
In this paper, we consistently employ the mean shape of the Basel Face Model as the generic 3D face model~\cite{paysan20093d}.
Given a 2D face image, it is first aligned to the generic 3D face model using orthogonal projection with the help of five facial feature points.
Then, the pre-labeled 3D landmarks are projected to the 2D image.
Lastly, a patch of size $100\times100$ pixels is cropped centering around each of the projected 2D landmarks.
More examples of the detected 2D uniform landmarks are shown in Fig.~\ref{fig:ThreeDProjectedLandmarks}.
It is clear that the patches are indeed uniformly sampled in the semantic meaning regardless of the pose variations of the face image.

\subsection{Combination of CNNs using Stacked Auto-Encoder}
We denote the features extracted by the set of CNNs as $\left\{x_{1},x_{2},\cdots,x_{K}\right\}$, where $x_{i}\in\mathbb R^{d\times 1}, 1\leq i\leq K$.
In this paper, $K$ equals to 8 and $d$ equals to 512.
The set of features represents multimodal information for face recognition.
We conduct feature-level fusion to obtain a single signature for each face image.
In detail, the features extracted by the eight CNNs are concatenated as a large feature vector, denoted as:
\begin{equation}
\hat{x}=[x_{1};x_{2};\cdots;x_{K}]\in\mathbb R^{Kd\times 1}.
\end{equation}

$\hat{x}$ is high dimensional, which is impractical for real-world face recognition applications.
We further propose to reduce the dimension of $\hat{x}$ by SAE.
Compared with the traditional dimension reduction approaches, \eg, PCA, SAE has advantage in learning non-linear feature transformations.
In this paper, we employ a three-layer SAE.
The number of the neurons of the three auto-encoders are 2048, 1024, and 512, respectively.
The output of the last encoder is utilized as the compact signature of the face image.
The structure for the designed SAE is illustrated in Fig.~\ref{fig:MMDFR_Framework}.

Nonlinear activation function is utilized after each of the fully-connected layers.
Two activation functions, \ie, sigmoid function and hyperbolic tangent (tanh) function, are evaluated.
The forward function of the sigmoid activation function is represented as
\begin{equation}
\begin{array}{cl}
S(x_{i})=\frac{1}{1+exp({-W_{f}^{T}x_{i}-b_{f}})}.
\end{array}
\label{E:Sigmoid}
\end{equation}
The forward function of the tanh activation function is represented as
\begin{equation}
\begin{array}{cl}
T(x_{i})=\frac{exp({W_{f}^{T}x_{i}+b_{f}})-exp({-W_{f}^{T}x_{i}-b_{f}})}{exp({W_{f}^{T}x_{i}+b_{f}})+exp({-W_{f}^{T}x_{i}-b_{f}})},
\end{array}
\label{E:TanH}
\end{equation}
where $x_{i}$, $W_{f}$, and $b_{f}$ are the input, weight, and bias of the corresponding fully-connected layer before the activation function.
Different normalization schemes of $\hat{x}$ are adopted for the sigmoid and tanh activation functions, since their output space is different.
For the sigmoid function, we normalize the elements of $\hat{x}$ to be within $\left[0,1\right]$.
For the tanh function, we normalize the elements of $\hat{x}$ to be within $\left[-1,+1\right]$.
In the experimentation section, we empirically compare the performance of SAE with the two different nonlinearities.

\section{Face Matching with MM-DFR}
In this section, the face matching problem is addressed based on the proposed MM-DFR framework.
Two evaluation modes are adopted: the unsupervised mode and the supervised mode.
Suppose two features produced by MM-DFR for two images are denoted as $y_{1}$ and $y_{2}$, respectively.
In the unsupervised mode, the cosine distance is employed to measure the similarity $s$ between $y_{1}$ and $y_{2}$.
\begin{equation}
s(y_{1},y_{2})=\frac{y_{1}^Ty_{2}}{\|y_{1}\|\|y_{2}\|}.
\end{equation}

For the supervised mode, a number of discriminative or generative models can be employed~\cite{prince2007probabilistic,chen2012bayesian,liu2015Classification},
In this paper, we employ the Joint Bayesian (JB) model~\cite{chen2012bayesian} as it is shown to outperform other popular models in recent works~\cite{ding2014multi}.
For both the unsupervised and supervised modes, the nearest neighbor (NN) classifier is adopted for face identification.
JB models the face generation process as
\begin{equation}
{x} = \mu  + {\varepsilon},
\end{equation}
where $\mu$ represents the identity of the subject, while $\varepsilon$ represents intra-personal noise.

JB solves the face identification or verification problems by computing the log-likelihood ratio
between the probability $P(x_{1},x_{2}|H_{I})$ that two faces belong to the same subject
and the probability $P(x_{1},x_{2}|H_{E})$ that two faces belong to different subjects.
\begin{equation}
{r(x_1,x_2)} = log\frac{P(x_{1},x_{2}|H_{I})}{P(x_{1},x_{2}|H_{E})},
\end{equation}
where $r(x_1,x_2)$ represents the log-likelihood ratio,
and we refer to $r(x_1,x_2)$ as similarity score for clarity in the experimental part of the paper.

\section{EXPERIMENTAL EVALUATION}
In this section, extensive experiments are conducted to present the effectiveness of the proposed MM-DFR framework.
The experiments are conducted on two large-scale unconstrained face databases, \ie, LFW~\cite{LFWTech} and CASIA-WebFace~\cite{yi2014learning}.
Images in both databases are collected from internet; therefore they are real images that appear in multimedia circumstances.

The LFW~\cite{LFWTech} database contains 13,233 images of 5,749 subjects.
Images in this database exhibit rich intra-personal variations of pose, illumination, and expression.
It has been extensively studied for the research of unconstrained face recognition in recent years.
Images in LFW are organized into two ``Views''. View 1 is for model selection and parameter tuning while View 2 is for performance reporting.
In this paper, we follow the official protocol of LFW and
report the mean verification accuracy and the standard error of the mean ($S_E$) by the 10-fold cross-validation scheme on the View 2 data.

Despite of its popularity, the LFW database contains limited number of images and subjects, which restricts its evaluation for large-scale unconstrained face recognition applications.
The CASIA-WebFace~\cite{yi2014learning} database has been released recently. CASIA-WebFace contains 494,414 images of 10,575 subjects.
As images in this database are collected in a semi-automatic way, there is a small amount of mis-labeled images in this database.
Because there is no officially defined protocol for face recognition on this database,
we define our own protocol for face identification in this paper.
In brief, we divide CASIA-WebFace into two sets: a training set and a testing set.
The 10,575 subjects are ranked in the descent order by the number of their images contained in the database.
The 471,592 images of the top 9,000 subjects compose the training set.
The 22,822 images of the rest 1,575 subjects make up the testing set.

\begin{figure}
\centering
\includegraphics[width=0.95\linewidth]{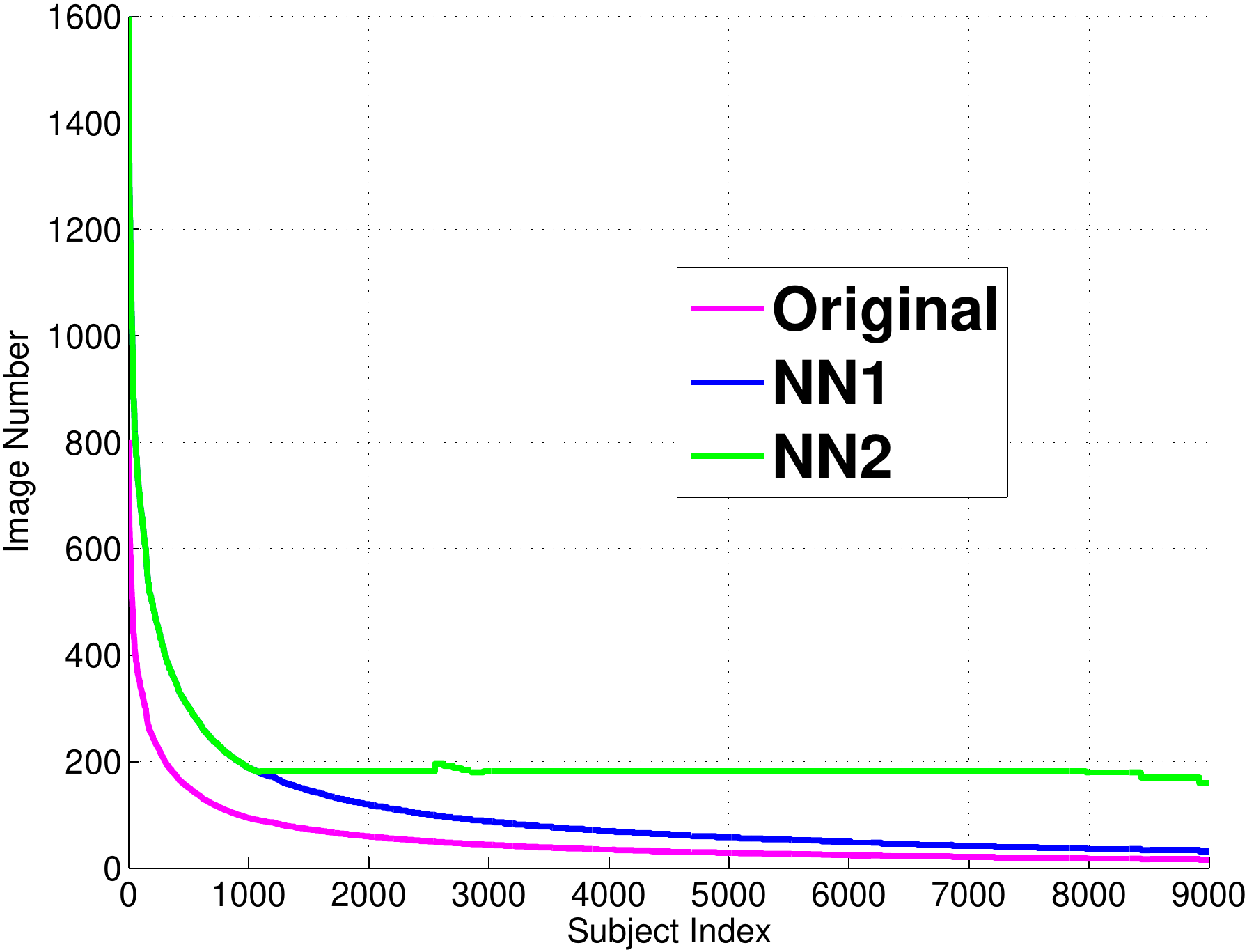}
\caption{Training data distribution for NN1 and NN2. This figure plots the number of images for each subject in the training set.
The long-tail distribution characteristic~\cite{zhou2015naive} of the original training data is improved after the aggressive data augmentation for NN2.}
\label{fig:DataDistribution}
\end{figure}

All CNNs and SAE in this paper are trained using the 9,000 subjects in the defined training set above.
Images are converted to gray-scale and geometrically normalized as described in Section III.
For NN1, we double the size of the training set by flipping all training images horizontally to reduce overfitting.
Therefore, the size of training data for NN1 is 943,184.
For NN2, we adopt much more aggressive data augmentation by horizontal flipping, image jittering\footnote{For image jittering,
we add random gaussian noise on the coordinates of the five facial feature points.
The noise is distributed with zero mean and standard deviation of four pixels.}, and image down-sampling.
The size of the augmented training data for NN2 is about 1.8 million.
The distribution of training data for NN1 and NN2 is illustrated in Fig.~\ref{fig:DataDistribution}.
It is shown that the long-tail distribution characteristic~\cite{zhou2015naive} of the original training data is improved after the aggressive data augmentation for NN2.

We adopt the following multi-stage training strategy to train all the CNN models.
First, we train the CNN models as a multi-class classification problem, \ie, softmax loss is employed.
For all CNNs, the initial learning rate for all learning layers is set to 0.01,
and is divided by 10 after 10 epochs, to the final rate of 0.001.
Second, we adopt the recently proposed triplet loss~\cite{schroff2015facenet} for fine-tuning for 2 more epochs.
We set the margin for the triplet loss to be 0.2 and learning rate to be 0.001.
It is expected that this multi-stage training strategy can boost performance while converge faster than using the triplet loss alone~\cite{schroff2015facenet}.
For SAE, the learning rate decreases from 0.01 to 0.00001, gradually.
We train each of the three auto-encoders one by one and each auto-encoder is trained for 10 epochs.
In the testing phase, we extract deep feature from both the original image and its horizontally flipped image.
Unless otherwise specified, the two feature vectors are averaged as the representation of the input face image.
The open-source deep learning toolkit Caffe~\cite{jia2014caffe} is utilized to train all the deep models.

Five sets of experiments are conducted.
First, we empirically justify the advantage of dense features for face recognition by excluding two ReLU nonlinearities compared with previous works.
The performance of the proposed single CNN model is also compared against the state-of-the-art CNN models on the LFW database.
Next, the performance of the eight CNNs contained within the MM-DFR framework is compared on face verification task on LFW.
Then, the fusion of the eight CNNs by SAE is conducted and different nonlinearities are also compared.
We also test the performance of MM-DFR followed with the supervised classifier JB.
Lastly, face identification experiment is conducted on the CASIA-WebFace database with our own defined evaluation protocol.

\subsection{Performance Comparison with Single CNN Model}
In this experiment, we evaluate the role of ReLU nonlinearity using CNN-H1 as an example.
For fast evaluation, the comparison is conducted with the simple NN1 structure described in Table~\ref{tab:NN1}
and only the softmax loss is employed for model training.
Performance of CNN-H1 using the NN2 structure can be found in Table~\ref{tab:elevenCNN}.
Two paradigms\footnote{Similar to previous works~\cite{taigman2014deepface,yi2014learning}, both the two paradigms defined in this paper
correspond to the ``Unrestricted, Labeled Outside Data Results'' protocol that is officially defined in~\cite{LFWTech}.} are followed:
1) the unsupervised paradigm that directly calculate the similarity between two CNN features using cosine distance metric.
2) the supervised paradigm that uses JB to calculate the similarity between two CNN features.
For the supervised paradigm, we concatenate the CNN features of the original face image and its horizontally flipped version as the raw representation of each test sample.
Then, we adopt PCA for dimension reduction and JB for similarity calculation.
The dimension of the PCA subspace is tuned on the View 1 data of LFW and applied to the View 2 data.
Both PCA and JB are trained on the CASIA-WebFace database.
For PCA, to boost performance, we also re-evaluate the mean of CNN features using the 9 training folds of LFW in 10-fold cross validation.

\begin{figure}
\centering
\includegraphics[width=0.9\linewidth]{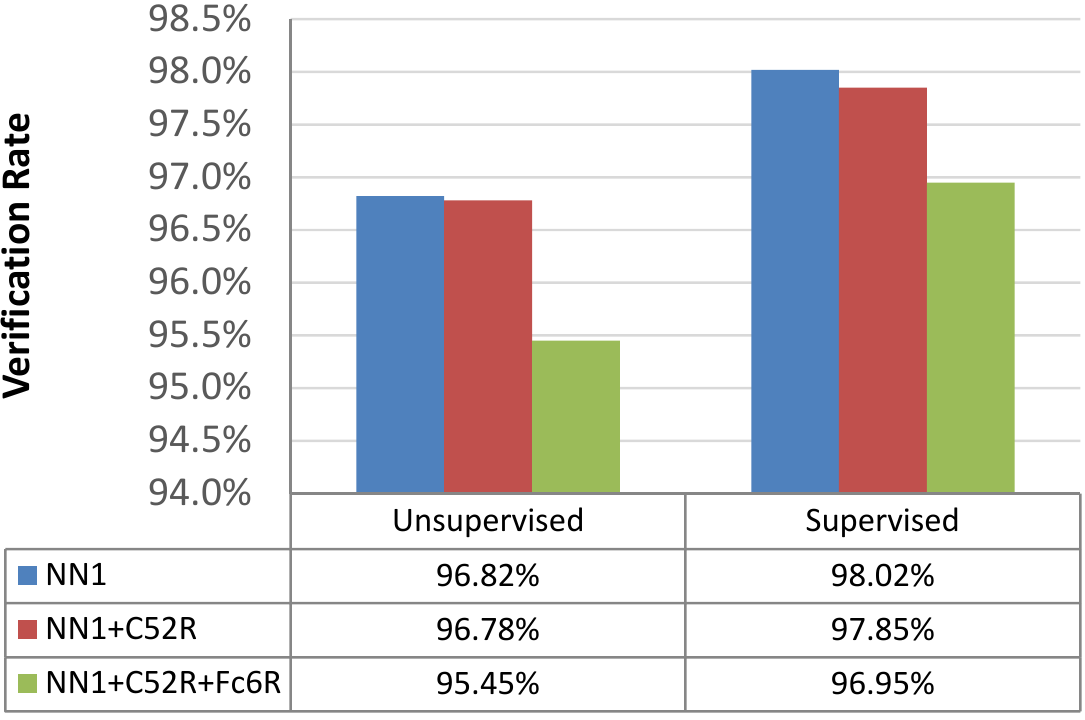}
\caption{Performance comparison on LFW with different usage strategies of ReLU nonlinearity.}
\label{fig:reluRoleTable}
\end{figure}

\begin{figure}
\centering
\includegraphics[width=1.00\linewidth]{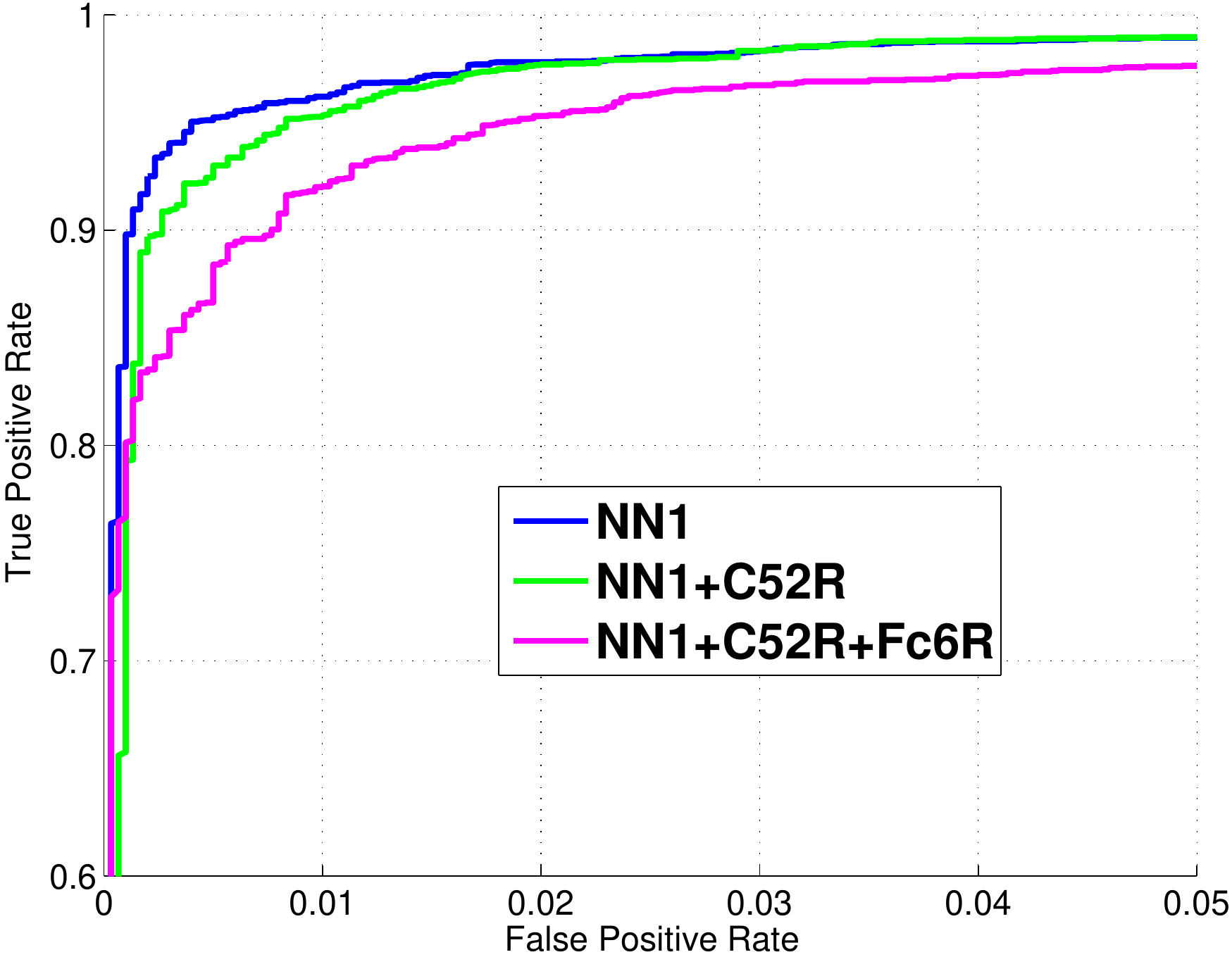}
\caption{ROC curves of different usage strategies of the ReLU nonlinearity on LFW.}
\label{fig:ReluStrategy}
\end{figure}

The performance of three structures are reported in Fig.~\ref{fig:reluRoleTable} and Fig.~\ref{fig:ReluStrategy}:
1) NN1, 2) NN1 with ReLU after Conv52 layer (denoted as NN1+C52R),
and 3) NN1 with ReLU after both Conv52 and Fc6 (denoted as NN1+C52R+Fc6R).
For both NN1+C52R and NN1+C52R+Fc6R, we replace the average pooling layer after Conv 52 with max pooling accordingly.
It is shown in Fig.~\ref{fig:reluRoleTable} that the ReLU nonlinearity after Conv52 or Fc6 actually harms the performance of CNN.
The experimental results have two implications: 1) dense feature is preferable than sparse feature for CNN, as intuitively advocated in~\cite{yi2014learning}.
However, there is no experimental justification in~\cite{yi2014learning}.
2) the linear projection from the output of the ultimate convolutional layer (Conv52) to the low-dimensional subspace (Fc6)
is better than the commonly adopted non-linear projection.
This is clear evidence that the negative response of the ultimate convolutional layer (Conv52) also contains useful information.

The performance by single CNN models on LFW is reported in Table.~\ref{tab:singleCNNComparison}.
The performance of the state-of-the-art CNN models is also tabulated.
Compared with Fig.~\ref{fig:reluRoleTable}, we further improve the performance of NN1 by fine-tuning with triplet loss.
It seems that the triplet loss mainly improves the performance for the unsupervised mode in our experiment.
It is shown that the proposed CNN model consistently outperforms the state-of-the-art CNN models under both the unsupervised paradigm and supervised paradigm.
In particular, compared with~\cite{yi2014learning,wang2015face} that all employ the complete CASIA-WebFace database for CNN training,
we only leverage a subset of the CASIA-WebFace database.
With more training data, we expect the proposed CNN model can outperform the other models with an even larger margin.

\begin{table}[!t]
\renewcommand{\arraystretch}{1.3}
\caption{Performance Comparison on LFW using Single CNN Model on Holistic Face Image}
\label{tab:singleCNNComparison}
\centering
\begin{tabular}{|l|c|c|}
\hline
                                            &Accuracy (Unsupervised)   &Accuracy (Supervised)\\ \hline\hline
DeepFace~\cite{taigman2014deepface}         &95.92 $\pm$ 0.29          &97.00 $\pm$ 0.87\\\hline
DeepID2~\cite{sun2014deep}                  &-                         &96.33 $\pm$ -\\\hline
Arxiv2014~\cite{yi2014learning}             &96.13 $\pm$ 0.30          &97.73 $\pm$ 0.31\\\hline
Facebook~\cite{taigman2014web}              &-                         &98.00 $\pm$ -\\\hline
MSU TR~\cite{wang2015face}                  &96.95 $\pm$ 1.02          &97.45 $\pm$ 0.99\\\hline
\textbf{Ours (NN1)}                         &\textbf{97.32 $\pm$ 0.34} &\textbf{98.05 $\pm$ 0.22}\\\hline
\textbf{Ours (NN2)}                         &\textbf{98.12 $\pm$ 0.24} &\textbf{98.43 $\pm$ 0.20}\\\hline
\end{tabular}
\end{table}

\subsection{Performance of the Eight CNNs in MM-DFR}
In this experiment, we present in Table~\ref{tab:elevenCNN} the performance achieved by each of the eight CNNs contained within the MM-DFR framework.
We report the performance of CNN-H1 with the NN2 structure while the other seven CNNs all employ the more efficient NN1 structure.
The same as the previous experiment, both the unsupervised paradigm and supervised paradigm are followed.
For the supervised paradigm, the PCA subspace dimension of the eight CNNs is unified to be 110.
Besides, features of the original face image and the horizontally flipped version are L2 normalized before concatenation.
We find that this normalization operation typically boosts the performance of the supervised paradigm by 0.1\% to 0.4\%.

\begin{table}[!t]
\renewcommand{\arraystretch}{1.3}
\caption{Performance Comparison on LFW of Eight Individual CNNs}
\label{tab:elevenCNN}
\centering
\begin{tabular}{|l|c|c|}
\hline
              &Accuracy (Unsupervised)   &Accuracy (Supervised)\\ \hline\hline
\textbf{CNN-H1}  &\textbf{98.12 $\pm$ 0.24}  &\textbf{98.43 $\pm$ 0.20}\\\hline
CNN-H2        &96.47 $\pm$ 0.44         &97.67 $\pm$ 0.28\\\hline
CNN-P1        &96.83 $\pm$ 0.26         &97.30 $\pm$ 0.22\\\hline
CNN-P2        &97.25 $\pm$ 0.31         &98.00 $\pm$ 0.24\\\hline
CNN-P3        &96.70 $\pm$ 0.25         &97.82 $\pm$ 0.16\\\hline
CNN-P4        &96.17 $\pm$ 0.31         &96.93 $\pm$ 0.21\\\hline
CNN-P5        &96.05 $\pm$ 0.27         &97.23 $\pm$ 0.20\\\hline
CNN-P6        &95.58 $\pm$ 0.17         &96.72 $\pm$ 0.21\\\hline
\end{tabular}
\end{table}

When combining Table~\ref{tab:singleCNNComparison} and Table~\ref{tab:elevenCNN},
it is clear that CNN-H1 outperforms CNN-H2 with the same NN1 structure, although they both extract features from holistic face images.
This maybe counter-intuitive, since the impact of pose variation has been reduced for CNN-H2.
We explain this phenomenon from the following two aspects:
1) most images in LFW are near-frontal face images, so the 3D pose normalization employed by CNN-H2 does not contribute much to pose correction.
2) the errors in pose normalization bring about undesirable distortions and artifacts to facial texture,
\eg, the distorted eyes, nose, and mouth shown in Fig.~\ref{fig:CroppedIm}(a).
The distorted facial texture is adverse to face recognition, as argued in our previous work~\cite{ding2015comprehensive}.
However, we empirically observe that the performance of MM-DFR drops slightly on View 1 data if we exclude CNN-H2,
which indicates CNN-H2 provides complementary information to CNN-H1 from a novel modality.
The contribution of CNN-H2 to MM-DFR is also justified by the last experiment in this section.
Besides, the performance of the patch-level CNNs, \ie, CNN-P1 to CNN-P6, fluctuates according to the discriminative power of the corresponding patches.

\subsection{Fusion of CNNs with SAE}
In this experiment, we empirically choose the best nonlinearity for SAE that is employed for feature-level fusion of the eight CNNs.
The structure of SAE employed in this paper is described in Fig.~\ref{fig:MMDFR_Framework}.
For each CNN, we average the features of the original image and the horizontally flipped version.
L2 normalization is conducted for each averaged feature before concatenating the features produced by the eight CNNs.
Similar to the previous experiment, we find this normalization operation promotes the performance of MM-DFR.
The dimension of the input for SAE is 4,096.
Two types of non-linearities are evaluated, the sigmoid non-linearity and the tanh non-linearity,
denoted as SAE-SIG and SAE-TANH, respectively.
The output of the third encoder (before the nonlinear layer) is utilized as the signature of the face image.
Cosine distance is employed to evaluate the similarity between two face images.
SAE are trained on the training set of CASIA-WebFace, using feature vectors extracted from both the original images and the horizontally flipped ones.
The performance of SAE-SIG and SAE-TANH is 98.33\% and 97.90\% on the View1 data of LFW, respectively.

SAE-TANH considerably outperforms SAE-SIG.
One important difference between the sigmoid non-linearity and the tanh non-linearity is that they normalize the elements of the feature to be within $\left[0,1\right]$
and $\left[-1,1\right]$, respectively.
Compared with the tanh non-linearity, the sigmoid non-linearity loses the sign information of feature elements.
However, the sign information is valuable for discriminative power.

\subsection{Performance of MM-DFR with Joint Bayesian}
The above three experiments have justified the advantage of the proposed CNN structures.
In this experiment, we further promote the performance of the proposed framework.

\begin{figure}
\centering
\includegraphics[width=0.95\linewidth]{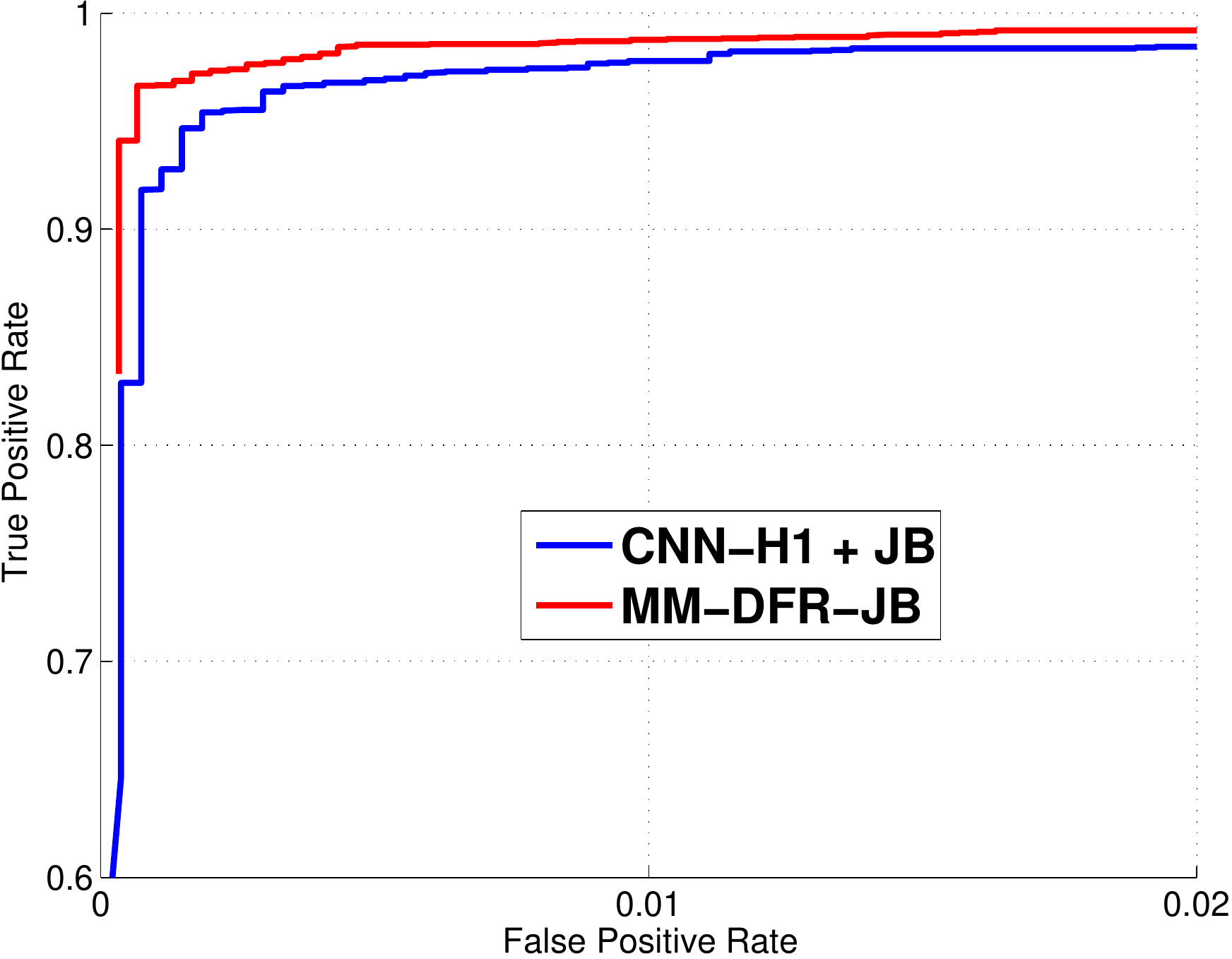}
\caption{Performance comparison between the proposed MM-DFR approach and single modality-based CNN on the face verification task.}
\label{fig:FusionVsSingle}
\end{figure}

We show the performance of MM-DFR with JB,
where the output of MM-DFR is utilized as the signature of the face image.
We term this face recognition pipeline as MM-DFR-JB.
For comparison, the performance achieved by CNN-H1 with the JB classifier is also presented, denoted as ``CNN-H1 + JB''.
The performance of the two systems is tabulated in Table~\ref{tab:supervisedClassCASIAWebFace} and the ROC curves are illustrated in Fig.~\ref{fig:FusionVsSingle}.
It is shown that MM-DFR considerably outperforms the single modal-based approach,
which indicates the fusion of multimodal information is important to promote the performance of face recognition systems.
By excluding the five labeling errors in LFW, the actual performance of MM-DFR-JB reaches 99.10\%.

Our simple 8-net based ensemble system also outperforms DeepID2~\cite{sun2014deep}, which includes as much as 25 CNNs.
Some more recent approaches that were published after the submission of this paper, \eg~\cite{schroff2015facenet,sun2015deeply}, achieve better performance than MM-DFR.
However, they either employ significantly larger private training dataset or considerably larger number of CNN models.
In comparison, we employ only 8 nets and train the models using a relatively small training set.

\begin{table}[!t]
\renewcommand{\arraystretch}{1.3}
\caption{Performance Evaluation of MM-DFR with JB}
\label{tab:supervisedClassCASIAWebFace}
\centering
\begin{tabular}{|l|c|c|}
\hline
                                        &\#Nets  &Accuracy(\%)$\pm S_E$\\ \hline\hline
CNN-H1 + JB                             &1       &98.43 $\pm$ 0.20\\\hline
DeepFace~\cite{taigman2014deepface}     &7       &97.25 $\pm$ 0.81\\\hline
MSU TR~\cite{wang2015face}              &7       &98.23 $\pm$ 0.68\\\hline
DeepID2~\cite{sun2014deep}              &25      &98.97 $\pm$ 0.25\\\hline
\textbf{MM-DFR-JB}                      &8       &\textbf{99.02 $\pm$ 0.19}\\\hline
\end{tabular}
\end{table}

\subsection{Face Identification on CASIA-WebFace Database}
The face identification experiment is conducted on the test data of the CASIA-WebFace database,
which includes 22,822 images of 1,575 subjects.
For each subject, the first five images are selected to make up the gallery set,
which can generally be satisfied in many multimedia applications, \eg, social networks where each subject has multiple face images.
All the other images compose the probe set.
Therefore, there are 7,875 gallery images and 14,947 probe images in total.

\begin{figure}
\centering
\includegraphics[width=1.00\linewidth]{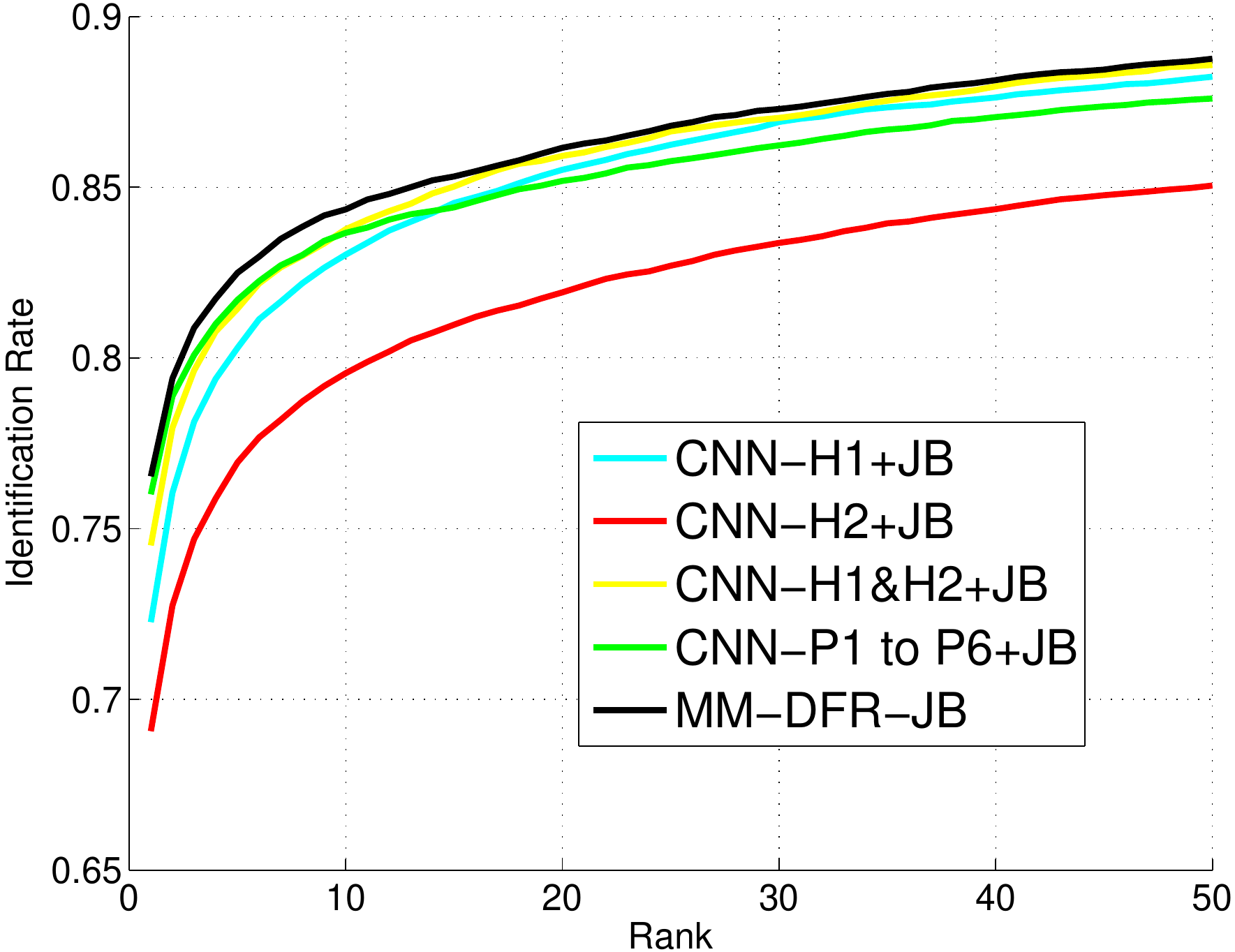}
\caption{CMS curves by different combinations of modalities on the face identification task.}
\label{fig:IdentificationCurve}
\end{figure}

The rank-1 identification rates by different combinations of modalities are tabulated in Table~\ref{tab:identificatonCASIAWebFace}.
The corresponding Cumulative Match Score (CMS) curves are illustrated in Fig.~\ref{fig:IdentificationCurve}.
It is shown that although very high face verification rate has been achieved on the LFW database,
large-scale face identification in real-world applications is still a very hard problem.
In particular, the rank-1 identification rate by the proposed approach is only 76.53\%.

It is clear that the proposed multimodal face recognition algorithm significantly outperforms the single modal based approach.
In particular, the rank-1 identification rate of MM-DFR-JB is higher than that of ``CNN-H1 + JB'' by as much as 4.27\%.
``CNN-H1 + JB'' outperforms ``CNN-H2 + JB'' with a large margin,
partially because CNN-H1 is based on the larger architecture NN2 and trained with more aggressively augmented data.
However, the combination of the two modalities still considerably boosts the performance by 2.25\% on the basis of CNN-H1,
which forcefully justifies the contribution of the new modality introduced by 3D pose normalization.
These experimental results are consistent with those obversed on the LFW database.
Experimental results on both datasets strongly justify the effectiveness of the proposed MM-DFR framework for multimedia applications.

\begin{table}[!t]
\renewcommand{\arraystretch}{1.3}
\caption{The rank-1 identification rates by Different Combinations of Modalities on CASIA-WebFace Database}
\label{tab:identificatonCASIAWebFace}
\centering
\begin{tabular}{|l|c|}
\hline
                       &Identification Rates\\ \hline\hline
CNN-H1 + JB            &72.26\%\\\hline
CNN-H2 + JB            &69.07\%\\\hline
CNN-H1\&H2 + JB        &74.51\%\\\hline
CNN-P1 to P6 + JB      &76.01\%\\\hline
MM-DFR-JB              &76.53\%\\\hline
\end{tabular}
\end{table}

\section{Conclusion}
Face recognition in multimedia applications is a challenging task because of the rich appearance change caused by pose, expression, and illumination variations.
We handle this problem by elaborately designing a deep architecture that employs complementary information from multimodal image data.
First, we enhance the recognition ability of each CNN by carefully integrating a number of published or our own developed tricks,
such as deep structures, small filters, careful use of ReLU nonlinearity, aggressive data augmentation, dropout, and multi-stage training with multiple losses, L2 normalization.
Second, we propose to extract multimodal information using a set of CNNs from the original holistic face image, the rendered frontal pose image by 3D model,
and uniformly sampled image patches.
Third, we present the feature-level fusion approach using stacked auto-encoders to fuse the features extracted by the set of CNNs,
which is advantageous to learn non-linear dimension reduction.
Extensive experiments have been conducted for both face verification and face identification experiments.
As the proposed MM-DFR approach effectively employs multimodal information for face recognition,
clear advantage of MM-DFR is shown compared with the single modal-based algorithms and some state-of-the-art deep models.
Other deep learning based approaches may also benefit from the structures that have been proved to be useful in this paper.
In the future, we will try to integrate more multimodal information into the MM-DFR framework and further promote the performance of single deep architecture such as NN2.


%



\section*{Acknowledgment}
The authors would like to thank the guest editor and the anonymous reviewers for their careful reading and valuable remarks.
This work is supported by Australian Research Council Projects FT-130101457 and DP-140102164.

\ifCLASSOPTIONcaptionsoff
  \newpage
\fi



\bibliographystyle{IEEEtran}


%




\end{document}